\documentclass[letterpaper, 10 pt, conference]{ieeeconf} 
\usepackage{url}
\usepackage{graphics}
\usepackage{graphicx}
\usepackage{caption}
\usepackage{subcaption}
\usepackage{algorithmic}
\usepackage{algorithm}
\usepackage{setspace}
\usepackage{epsfig}
\usepackage{amsfonts}
\usepackage{multirow}
\usepackage{makecell}
\usepackage{todonotes}
\usepackage{pbox}
\usepackage{soul}

\usepackage{array}
\usepackage{amssymb,url,amsmath}
\usepackage{algorithm, algorithmic}
\usepackage{graphicx}
\usepackage{color}
\usepackage{multirow}
\usepackage{balance}

\usepackage{booktabs}       
\usepackage{amsfonts}       
\usepackage{nicefrac}       
\usepackage{microtype}      
\usepackage{caption}
\usepackage{booktabs}
\usepackage{tikz}
\usepackage{array}
\usepackage{nicefrac}

\usepackage{color}
\definecolor{CommentRed}{rgb}{0.7,0,0}
\definecolor{CommentBlue}{rgb}{0,0,0.7}
\definecolor{CommentGreen}{rgb}{0,0.7,0}

\newcolumntype{L}[1]{>{\raggedright\let\newline\\\arraybackslash\hspace{0pt}}m{#1}}
\newcolumntype{C}[1]{>{\centering\let\newline\\\arraybackslash\hspace{0pt}}m{#1}}
\newcolumntype{R}[1]{>{\raggedleft\let\newline\\\arraybackslash\hspace{0pt}}m{#1}}

\newcommand{\method}{Incremental Adversarial Domain Adaptation}
\newcommand{\met}{IADA}
\newcommand{\metadd}{SDM}
\newcommand{\etal}{\textit{et al}. }

\newcommand{\eg}{\textit{e}.\textit{g}. }

\newcommand{\captionSize}{\normalsize}

\definecolor{myblue}{RGB}{0,51,102}
\IEEEoverridecommandlockouts                              
\title{\LARGE \bf \method \\ for Continually Changing Environments  }

\author{Markus Wulfmeier$^{1}$, Alex Bewley$^{1}$ and Ingmar Posner$^{1}$
\thanks{$^{1}$The authors are with the Applied Artificial Intelligence Lab, Oxford Robotics Institute, 
        University of Oxford, United Kingdom; \newline 
        {\tt\small markus, bewley, ingmar@robots.ox.ac.uk}}}%

\begin{document}

\maketitle
\thispagestyle{empty}
\pagestyle{empty}

\maketitle

\begin{abstract}

Continuous appearance shifts such as changes in weather and lighting conditions can impact the performance of deployed machine learning models. While unsupervised domain adaptation aims to address this challenge, current approaches do not utilise the continuity of the occurring shifts.
In particular, many robotics applications exhibit these conditions and thus facilitate the potential to incrementally adapt a learnt model over minor shifts which integrate to massive differences over time.
Our work presents an adversarial approach for lifelong, incremental domain adaptation which benefits from unsupervised alignment to a series of intermediate domains which successively diverge from the labelled source domain.
We empirically demonstrate that our incremental approach improves handling of large appearance changes, \eg  day to night, on a traversable-path segmentation task compared with a direct, single alignment step approach.
Furthermore, by approximating the feature distribution for the source domain with a generative adversarial network, the deployment module can be rendered fully independent of retaining potentially large amounts of the related source training data for only a minor reduction in performance.

\end{abstract}

\section{Introduction}
\label{sec:introduction}

Appearance changes based on lighting, seasonal, and weather conditions provide a significant challenge for outdoor robots relying on machine learning models for perception. 
While providing high performance in their training domain, visual shifts occurring in the environment can result in significant deviations from the training distribution, severely reducing accuracy during deployment.
Commonly, this challenge is partially counteracted by employing additional training methods to render these models invariant to their application domain \cite{ratnasingam2010study}.

For scenarios where labelled data is unavailable in the target domain, the problem can be addressed in the context of unsupervised domain adaptation \cite{Long0J16a,Ganin2016}. 
Recent state-of-the-art approaches which address this challenge operate by training deep neural networks within an adversarial domain adaptation (ADA) framework.
These approaches are characterised by the optimisation of potentially multiple encoders with the objective to confuse a domain discriminator operating on their output \cite{Ganin2016,ajakan2014domain, wulfmeier2017addressing} in additional to their main objective. 
The main intuition behind this framework is that by training the encoder to obtain a domain invariant embedding, we allow the main supervised task to be robust to changes in the application domain. 

\begin{figure}[t]
	\centering
		\includegraphics[width = 0.47\textwidth]{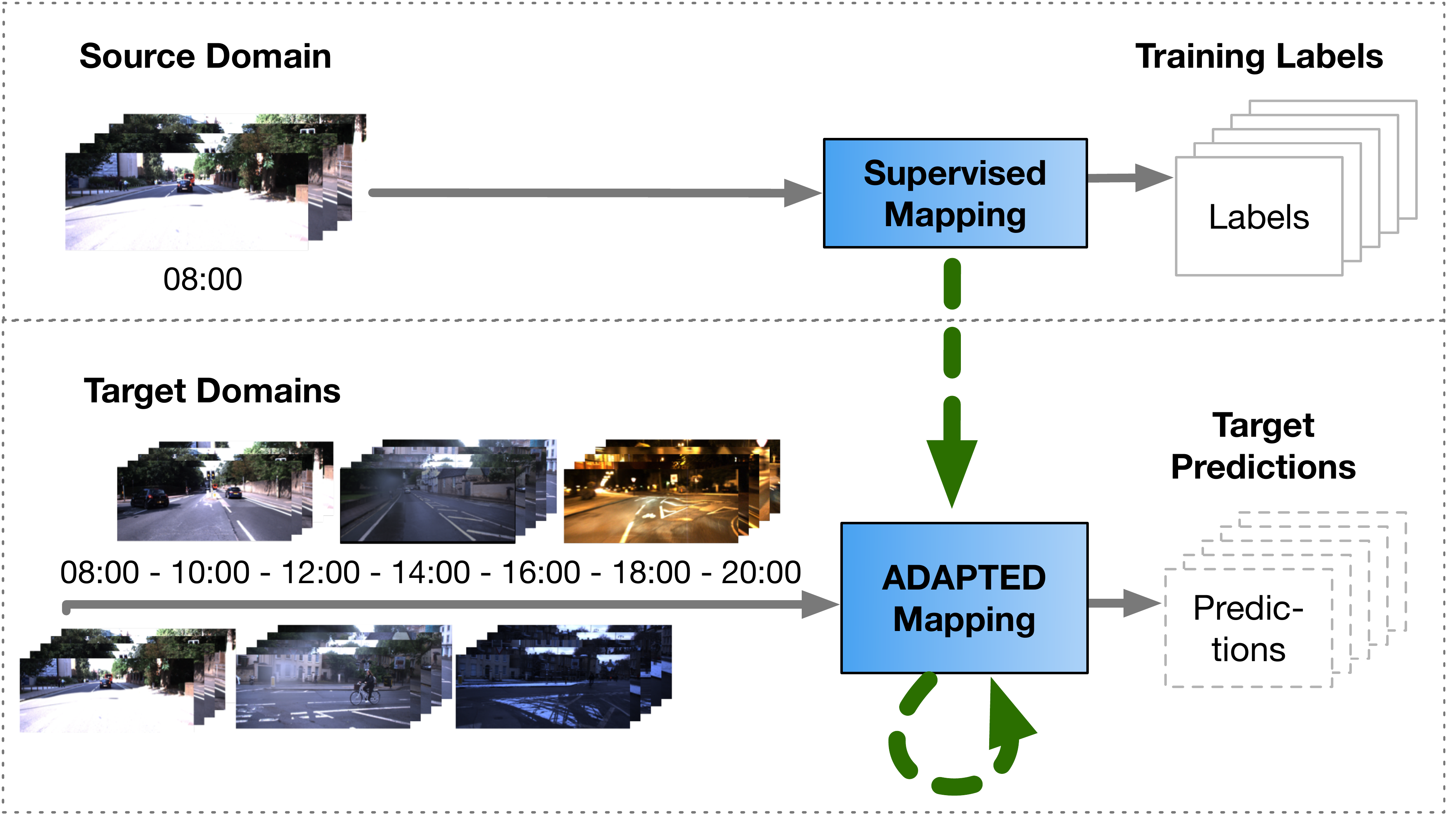}
    \caption{\captionSize \method. Instead of performing domain adaptation over large shifts at once, \met~ splits domain alignment into simpler subtasks. After adapting the feature embedding of the initial target domain, the approach incrementally refines all modules to the currently perceived target domains.}
	\label{fig:cada}
\end{figure}

Recent successes based on adversarial domain adaptation have achieved state-of-the-art performance on toy datasets \cite{Ganin2016,bousmalis2016domain, tzeng2015towards,tzeng2017adversarial} as well as real-world applications for autonomous driving within changing environmental conditions \cite{hoffman2016fcns,wulfmeier2017addressing}.
However, domains with a significant difference in appearance - such as day and night - continue to present a substantial challenge \cite{wulfmeier2017addressing}. 
We conjecture that the observed change in environmental conditions in many domains of continuous deployment (\eg in autonomous driving) is up to some extend composed via a gradual process which accumulates to produce massive differences over extended periods. 
This work exploits the incremental changes observed throughout deployment to continuously counteract the domain shift by updating discriminator and encoder incrementally while observing the visual shift (as illustrated in Figure \ref{fig:cada}).

During domain adaptation training, existing methods rely on data from the source domain to ensure training the discriminator based on a balanced distribution of source and target data to prevent overfitting to the target domain.
These approaches therefore require the storage of potentially massive datasets, which can provide a challenge in particular for mobile applications with limited available memory.

Similar to work on synthetic dataset extension \cite{ShrivastavaPTSW16}, we remove this requirement by training a generative adversarial network (GAN) \cite{goodfellow2014generative} to imitate the marginal encoder feature distribution in the source domain. We empirically demonstrate that domain adaptation via aligning target encoder features with GAN generated samples instead of source domain feature embeddings only results in minor performance reduction. Crucially, this means that the deployment module is fully independent of the size of source domain dataset, enabling application on mobile platforms with limited memory resources.

The approach is evaluated both on synthetic and real-world data. An artificial dataset is created with direct control over the number of intermediate domains and the strength of the incremental shifts for illustration and demonstration purposes. The following real-world evaluation focuses on a drivable-path segmentation task for autonomous driving based on segments from the Oxford RobotCar Dataset \cite{RobotCarDatasetIJRR} with different illumination conditions from different times of day.

The contributions of our work are as follows:
\begin{itemize}
	\item Introduction of a method for \emph{incremental} unsupervised domain adaptation for platforms deployed in continuously changing environments.
	\item Presentation of an additional method to remove the requirement of retaining extensive amounts of source data by modelling the feature representation of source domain data with a generative model.
	\item Quantitative investigation of the influence of dividing the adaptation task into incremental alignment over smaller shifts based on a synthetic toy example.
	\item Application of the proposed method to the real-world task of drivable-terrain segmentation and proof of feasibility for online application in the context of run-time evaluation on an NVIDIA GPU.
\end{itemize}

\section{Related Work}
\label{sec:related}


Continuously changing environment appearances have been a long-standing challenge for robot deployment as shifts between training and deployment data can seriously degrade model performance. 
Considerable efforts have been focused on designing and comparing various feature transformations with the goal of creating representations invariant to environmental change \cite{lowry2016visual}.
Other approaches address the problem through retaining multiple experiences \cite{ChurchillIJRR2013} or synthesising images between discrete domains \cite{neubert2013appearance}. However, it is unclear how these systems can efficiently scale to a continuous shift in the domain distribution.
 
In recent years, there has been a steady trend towards applying deep networks for various robotics tasks, where early layers act as a feature encoder with a supervised loss for the desired task on the output of the network.
Unfortunately, even such powerful models still suffer from the problem of shifts in domain appearance.
This has prompted a number of works which try to address this issue \cite{Ganin2016, hoffman2016fcns, Hong2017, wulfmeier2017addressing}.

The possibility to directly optimise complete feature representations via backpropagation for domain invariance \cite{Ganin2016} or target-source mappings \cite{Long0J16a} has lead to significant success of deep architectures in this field.
Long \etal \cite{Long0J16a} focus on minimising the Maximum Mean Discrepancy for the feature distributions of multiple layers of the network architecture. Rozantsev \etal \cite{rozantsev2016beyond} extend in a similar direction and impose a penalty for deviations in the network parameters across domains. Sun \etal \cite{SunFS16} align second order statistics of layer activations for source and target domains. Hoffman \etal \cite{hoffman2016fcns} match the label statistics between the true source and predicted target labels for semantic classification.

Furthermore, adversarial approaches to domain adaptation have been introduced~\cite{Ganin2016, ajakan2014domain, bousmalis2016domain}, which rely on training a domain discriminator to classify the domains underlying an encoder's feature distribution.
While adversarial training techniques have been shown to be notoriously unstable and difficult to optimise, there has been a pronounced body of work towards improving their stability, including more dominant use of the confusion loss \cite{goodfellow2014generative} and more recently the Wasserstein GAN framework \cite{arjovsky2017towards}. 

All the above mentioned works treat the unsupervised domain adaptation problem as a batch transition without exploiting temporal coherence commonly available to robots in continuous deployment. 
Continuous refinement has however been actively researched in supervised learning for many years (\eg \cite{kembhavi2009incremental,jain2011online, bewley2014online}), yet there has been little work on methods for unsupervised domain adaptation.
One notable exception is the work by Hoffman \etal \cite{hoffman2014continuous}, which addresses the problem with predefined features and focuses on the challenges of aligning to a continuously reshaping target domain.
This work seeks to extend the recently developed approach of adversarial domain adaption to a continuously evolving target domain by capitalising on the perpetual observations made by a robot.

\section{Method}
\label{sec:problem}

\method~addresses the problem of degraded model performance due to continuously shifting environmental conditions. This includes changes caused by weather and lighting occurring in outdoor scenarios. Compared to the regular single-step domain adaptation paradigm, we benefit in applications building on continual deployment through exploitation of the incremental changes that integrate to large domain shifts. Continuously observable lighting or seasonal shifts in outdoor robotics and other applications constitute a prime example for this paradigm.

\begin{figure}[t]
	\centering
		\includegraphics[width = 0.47\textwidth]{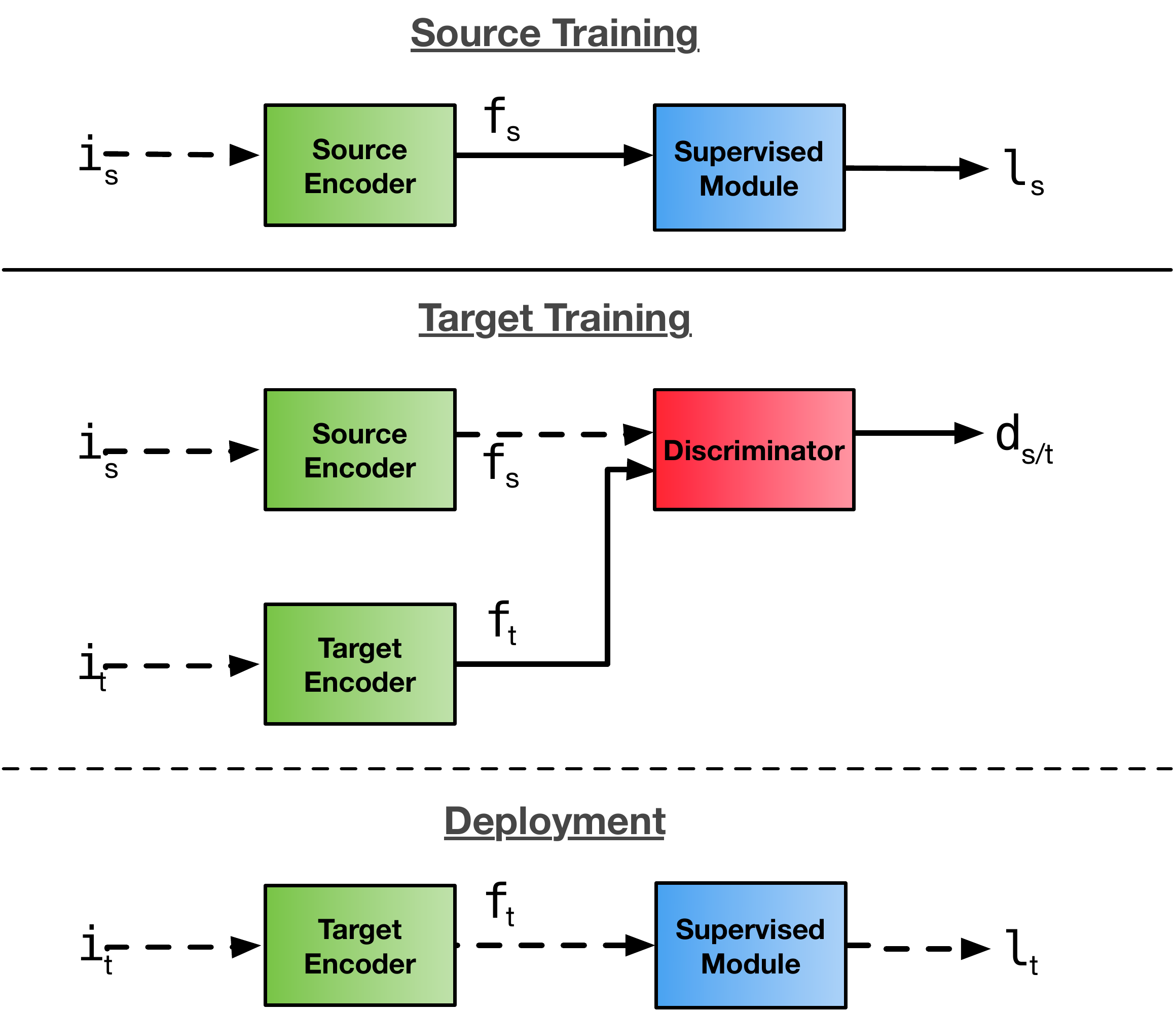}
    \caption{\captionSize Network architecture and information flow for \met. After the optimisation of source encoder and supervised model, the target encoder is trained to confuse the domain discriminator, leading to domain invariant feature representations. During deployment, the target encoder is connected to the supervised module. Dotted arrows represent only forward passes while solid lines display forward and gradient backward pass.}
	\label{fig:cada_network}
\end{figure}

The approach extends adversarial domain adaptation approaches \cite{Ganin2016} aiming to facilitate learning a feature encoding $f$ which is invariant with respect to the origin domain of its input data. In this way, the method enables the application of a supervised module trained only on source domain data to incoming unsupervised data from the application domain as depicted in Figure \ref{fig:cada_network}.

In comparison to existing methods \cite{Ganin2016,tzeng2015towards,wulfmeier2017addressing} which frame the task of unsupervised domain adaptation as a one-step approach between distinct source and target domains, \met~treats the incoming data as a stream of incrementally changing domains. 
Exploiting access to data from these incremental changes facilitates alignment over greater overall shifts between the target and source domains. The encoder and discriminator models are updated gradually to enable alignment for each incrementally shifted domain. 

Adversarial domain adaptation~\cite{Ganin2016} generally tends to be hyperparameter search intensive~\cite{wulfmeier2017addressing} as it - in addition to the adversarial min-max problem - is affected by the potential conflict of the domain invariance objective and the supervised objective. Intuitively, by dividing the domain alignment procedure into smaller incremental shifts, we simplify the overall task which can minimise the loss of relevant information. 

The training procedure is split into two principal segments: \textit{offline} supervised optimisation on source domain data and the unsupervised domain adaptation procedure, which potentially can be run \textit{online} during platform deployment as displayed in Figure \ref{fig:cada_network}. 

Hereinafter, let $\theta_X$ be the parametrisation of module $X$ and $i$ the incoming images. Source and target domains are represented with subscripts $s$ and $t$ respectively. The supervised training procedure optimises the supervised module $S$ with the predicted label $l=S(f,\theta_{S})$  as well as the source domain encoder $E_s$ based on a supervised task (e.g. classification or segmentation as in Section \ref{sec:experiments}). 

The parameters of both of these modules remain unchanged during training for domain adaptation, which enables us to keep source performance unaffected (an approach suggested for regular ADA in \cite{tzeng2017adversarial}). Only the target encoder $E_t$ and discriminator $D$ are trained via their respective objectives $\mathcal{L}_{E_t}$ and $\mathcal{L}_{D}$ in Equations \ref{eq:enc_loss} and \ref{eq:discr_loss} to align the target and source encoder feature spaces. 
Let $f_s = E_s(i_s,\theta_{E_s})$ and $f_t = E_t(i_t,\theta_{E_t})$ respectively denote the feature encoding of source and target images $i_s$ and $i_t$.

\begin{eqnarray}\label{eq:da_loss}
    \mathcal{L}_{E_t}( \theta_{E_t},\theta_{D}) &=& - \mathbb{E}_{i_t\sim T}[\log(D(f_t,\theta_{D}))]\label{eq:enc_loss}\\
    \mathcal{L}_{D}(\theta_{E_s},\theta_{E_t}, \theta_D) &=& -\mathbb{E}_{i_s\sim S }[ \log(D(f_s,\theta_D))]  \label{eq:discr_loss}\\
    \nonumber &&-\mathbb{E}_{i_t\sim T}[ \log(1 - D(f_t,\theta_D))]
\end{eqnarray}


The target encoder weights are initialised with parameters from the source encoder trained on the supervised task. These inchoate parameters are then subsequently adapted to align to the currently encountered target data by optimising both the target encoder and discriminator using the objectives in Equations \ref{eq:enc_loss} and \ref{eq:discr_loss} respectively. Intuitively, this procedure entails using the optimised parameters from the previously encountered target domain as initialisation for adapting to the current domain. 
The currently encountered unsupervised data is hereby utilised to fill a buffer from which is continuously sampled for the domain adaption training procedure.

%

\subsection{Source Distribution Modelling}
\label{sec:sdm}

\begin{figure}[t]
	\centering
		\includegraphics[width = 0.47\textwidth]{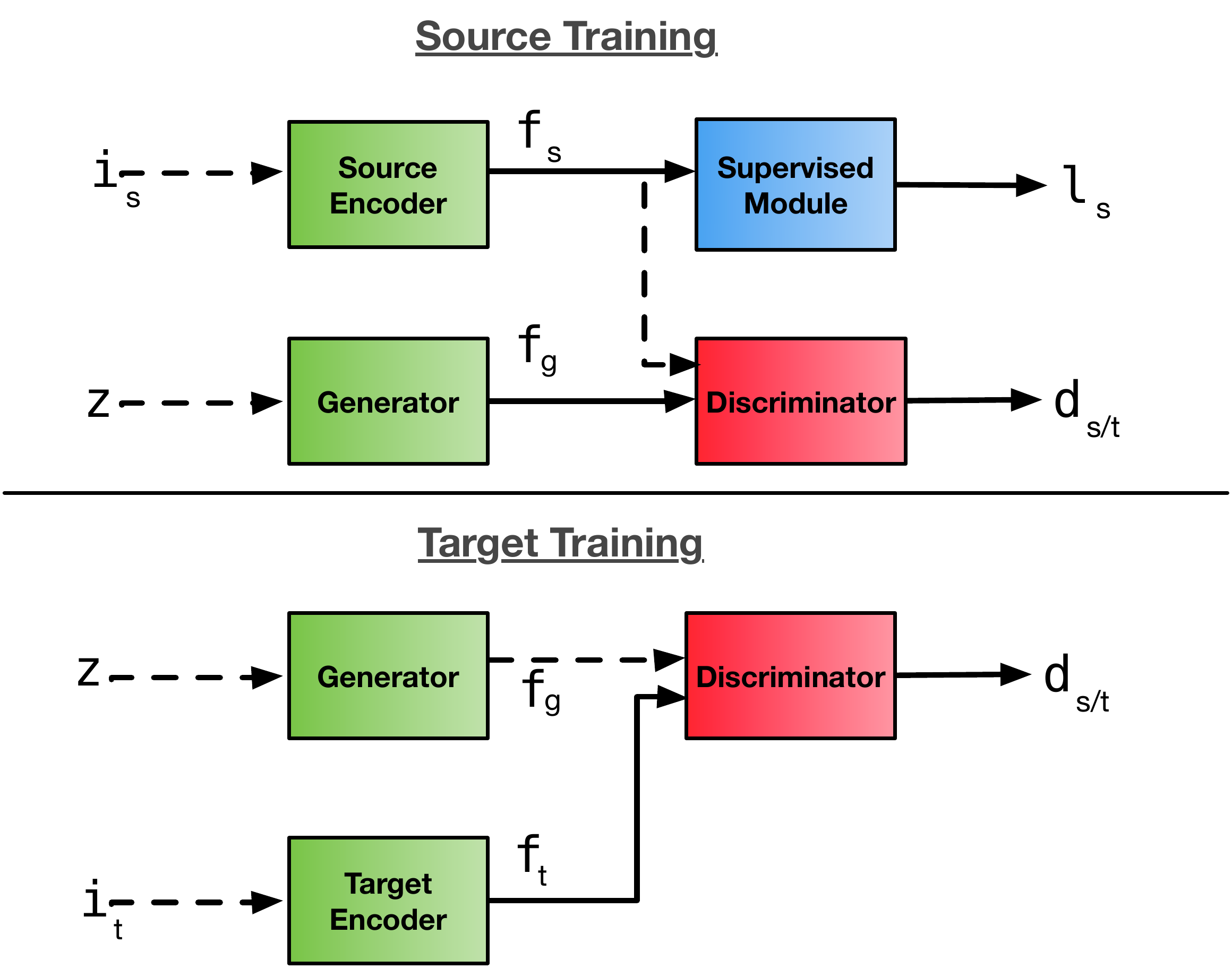}
    \caption{\captionSize Network architecture and information flow for training with a generative model approximating the marginal source feature distribution. The approach additionally trains a GAN during the source training procedure but does not propagate gradients for the adversarial loss to the source encoder to ensure unmodified source domain performance. Subsequently the target encoder is trained to mimic the feature distribution of the - now fixed - GAN. Dotted arrows represent only forward passes while solid lines display forward and gradient backward pass.}
	\label{fig:idasda}
\end{figure}

As \met's benefits apply in the context of continuously deployed platforms, the inherent requirement of ADA-based methods to retain potentially large amounts of source training data can constrain its application. Limitations on the weight and size of platforms often leads to restrictions on computational resources including data storage. 
To counteract this requirement, we additionally extend our method with a GAN-based approach to mimic the source domain's feature distribution, thus rendering the approach independent of the amount of source data during the domain adaptation task. 

More concretely, we optimise a generator $G$ which maps from n-dimensional, normally distributed noise $z\sim\mathcal{N}(\mu=0,\sigma=1)$ to approximate the feature distribution in our source domain during the offline training step. 
While the original GAN framework \cite{goodfellow2014generative} aims at mimicking natural images, our approach simply aims to imitate the feature encoding of images  (displayed in Figure \ref{fig:idasda}). The resulting objectives for generator $\mathcal{L}_{G}$ and discriminator $\mathcal{L}_{D}$ are displayed in Equations \ref{eq:g_loss_s} and \ref{eq:d_loss_s}. Let $f_g$ denote the generator features generated as $f_g=G(z,\theta_G)$.

\begin{eqnarray}
    \mathcal{L}_{G}( \theta_G, \theta_D) &=& - \mathbb{E}_{z\sim \mathcal{N}(\mu,\sigma) }[\log(D(f_g,\theta_D))]  \label{eq:g_loss_s}\\
    \mathcal{L}_{D}( \theta_G, \theta_{E_s}, \theta_D) &=& -\mathbb{E}_{i\sim S }[ \log(D(f_s,\theta_D))]  \label{eq:d_loss_s}\\
    \nonumber &&-\mathbb{E}_{z\sim \mathcal{N}(\mu,\sigma)}[ \log(1 - D(f_g,\theta_D))]
\end{eqnarray}


Subsequently during the domain adaptation procedure, the target encoder is optimised to align to the feature distribution of the GAN, whose parameters remain static to model the source domain. Instead of optimising the discriminator to classify between source and target domain, in this scenario it learns to distinguish between synthetically generated source features and actual target features encoding target images. Target encoder and discriminator are optimised towards the objectives $\mathcal{L}_{E_t}$ and $\mathcal{L}_{D}$ in Equations \ref{eq:e_loss_t} and \ref{eq:d_loss_t} respectively.

\begin{eqnarray}
    \mathcal{L}_{E_t}( \theta_{E_t}, \theta_D) &=& - \mathbb{E}_{i\sim T}[\log(D(f_t,\theta_{D}))]\label{eq:e_loss_t} \\
    \mathcal{L}_{D}( \theta_G,\theta_{E_t},\theta_D) &=& -\mathbb{E}_{z\sim \mathcal{N}(\mu,\sigma)}[ \log(D(f_g,\theta_D))] \label{eq:d_loss_t}\\
    \nonumber &&-\mathbb{E}_{i\sim T}[ \log(1 - D(f_t,\theta_D))]
\end{eqnarray}
\vspace{1mm}


Similar to \met~, we utilise all models, which are trained in the source domain, as initialisation for training in the target domain. For \metadd, this procedure additionally includes the discriminator. Lastly, the deployment setup is equivalent to \met.

\section{Experiments}
\label{sec:experiments}

Our evaluation is split into two parts: we first investigate a toy scenario with artificially induced domain shift for the purpose of visualisation and clarification, then we demonstrate performance gains in a continuous deployment scenario for drivable-path segmentation for autonomous mobility.

The evaluation compares \met~against its one-step counterpart ADA, and furthermore investigates the influence of source domain modelling based on Section \ref{sec:sdm}. The evaluation metric depends on the supervised task in the respective target domains, classification accuracy for the toy example and mean average precision for the drivable-path segmentation task.

While ADA only utilises the final source domain, \met~has access to all incremental domains. To evaluate if the cause of \met's advantages simply is the reliance on a larger dataset, we additionally introduce ADA Union. The method combines all target domains into a single dataset and performs regular ADA with respect to this union over all target domains.

\subsection{Toy Example: Incrementally Transformed MNIST}
\label{sec:mnist}

To quantify the benefits of \met~in relation to the strength of domain shifts and the number of intermediate domains, we first evaluate the approach in a scenario based on increasingly, synthetically deformed versions of the popular MNIST dataset.

\begin{figure}[t]
	\centering
		\includegraphics[width = 0.4\textwidth,trim={0cm 2cm 0 2cm},clip]{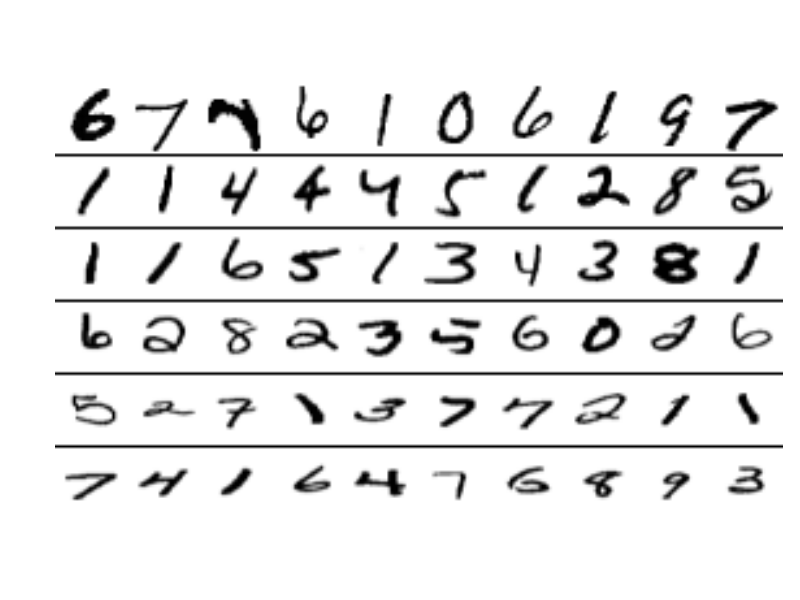}
    \caption{\captionSize Incremental deformation of MNIST digits from full to half height over 5 intermediate domains. Top row: original source data. Bottom row: maximally transformed target domain.}
	\label{fig:mnist}
\end{figure}

We create additional, transformed copies of the original dataset with height-rescaled digits of between 0.9 to 0.5 times the original height, which are visualised in Figure \ref{fig:mnist}. These synthetically transformed domains enable us to create a scenario with full control over the underlying domain shift and ensure that the occurring changes can be observed and utilised for domain adaptation in arbitrary detail.

We employ a Network-in-Network like architecture \cite{lin2013network} with exponential linear activation functions \cite{clevert2015fast} splitting after the last hidden layer and applying a discriminator with 2 hidden layers and each 512 neurons. 
All parameters before the split are duplicated for source and target encoders while the supervised module consists of the last fully connected layer. 
The adversarial loss is weighted by a factor of 0.001 for domain adaptation as well as training the GAN as part of the source domain modelling step. 

\begin{table}[ht]
    \centering
    \begin{tabular}{c|c|c|c|c|c|c}
       \toprule
            \makecell{target\\ domains} & \makecell{only\\ source} & ADA & \makecell{ADA \\\metadd} & \makecell{ADA\\Union} & \met & \makecell{\met~\\ \metadd} \\
        \midrule
         0.9 & 99.31 & - & - & - & 99.61 & 99.52 \\
         0.8 & 99.20 & - & - & - & 99.53 & 99.36 \\
         0.7 & 98.40 & - & - & - & 99.20 & 99.01 \\
         0.6 & 93.51 & - & - & - & 95.68 & 95.11 \\
         0.5 & 84.11 & 87.10 & 86.83 & 87.62 & 89.90 & 89.51 \\ 
        \bottomrule
    \end{tabular}
    \vspace{3pt}
    \caption{\captionSize Target classifier accuracy on incrementally transformed MNIST dataset. The last row represents the final accuracy on the maximally transformed input samples. While the naive alignment to the union of all target datasets already improves in performance of an approach with only access to the final domain, \met~results in further significant accuracy improvement. \metadd~only slightly affects the target performance and the combination \met~\metadd~continues to outperform the original ADA baseline.}
    \label{tab:mnist}
\end{table}

Table \ref{tab:mnist} shows the target domain classification accuracy of 1-step adaptation methods against their incremental counterparts which continue optimising the target encoder across domains with incrementally increasing domain shift. Furthermore, we test the methods in combination with GAN-based Source Domain Modelling (\metadd) introduced in Section \ref{sec:sdm}. 

As displayed in Table \ref{tab:mnist}, all domain adaptation approaches outperform pure source-optimised models, while incremental domain adaptation provides additional benefits over regular ADA. Utilising the union of all target domains as target domain for regular ADA increases model accuracy in the final target domain only slightly above using only one target domain and still performs worse than \met. Finally, the SDM variants of all approaches only result in minor performance reductions, while reducing memory requirements significantly.


To investigate classification accuracy in dependence of the number of available intermediate domains, the MNIST digits are rescaled further to 0.3 of the original height and evaluated with varying numbers of equally spread intermediate domains with \met~and \met~\metadd.  
We chose to increase the deformation in comparison to the earlier experiment to increase the complexity of the domain adaptation task and enable evaluating a wider range of sub-domain discretisations.


\begin{figure}[ht]
    \centering
    \includegraphics[width=.5\textwidth]{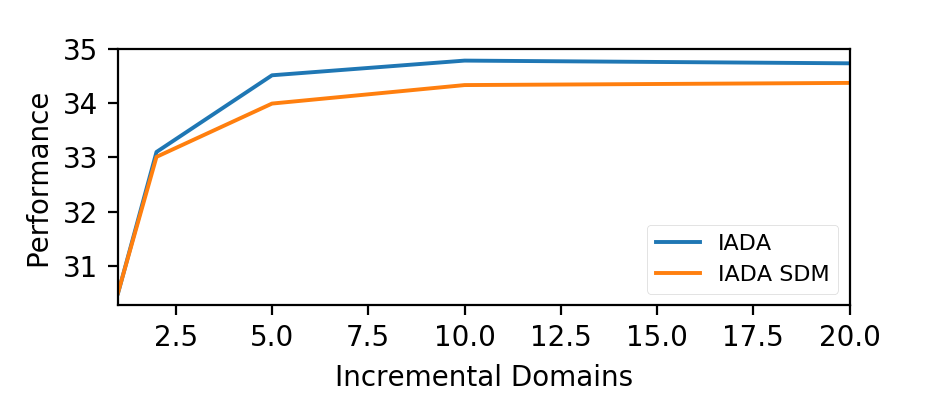}
    \caption{\captionSize Classifier accuracy of \met~in final target domain with varying number of intermediate domains for horizontal compression of 0.3. The strong digit deformation leads to a challenge for domain adaptation. Results show the benefits of separating large domain shifts into incremental domain adaptation steps for \met. Maximal performance for this adaptation scenario is achieved between 10 and 20 incremental domains and further increase does not significantly influence the final target accuracy.}
    \label{fig:subdomains}
\end{figure}

Separating larger shifts into incremental steps as displayed in Figure \ref{fig:subdomains} enables us to address the problem with a curriculum of easier tasks. Above a certain threshold however target performance remains consistent with further increase of the number of target domains.  
For the domain adaptation from MNIST to its rescaled copy, the benefits of incremental domain adaptation saturate at around 10 to 20 intermediate domains. However, more complex transformations can rely on even more incremental approaches.

\subsection{Free Space Segmentation}
\label{sec:segmentation}

An area of active research in autonomous driving is the detection of traversable terrain. Especially when utilising images as input, methods often rely on collecting data in all possible deployment domains and weather conditions. 

\begin{figure}[t]
	\centering
		\includegraphics[width = 0.48\textwidth]{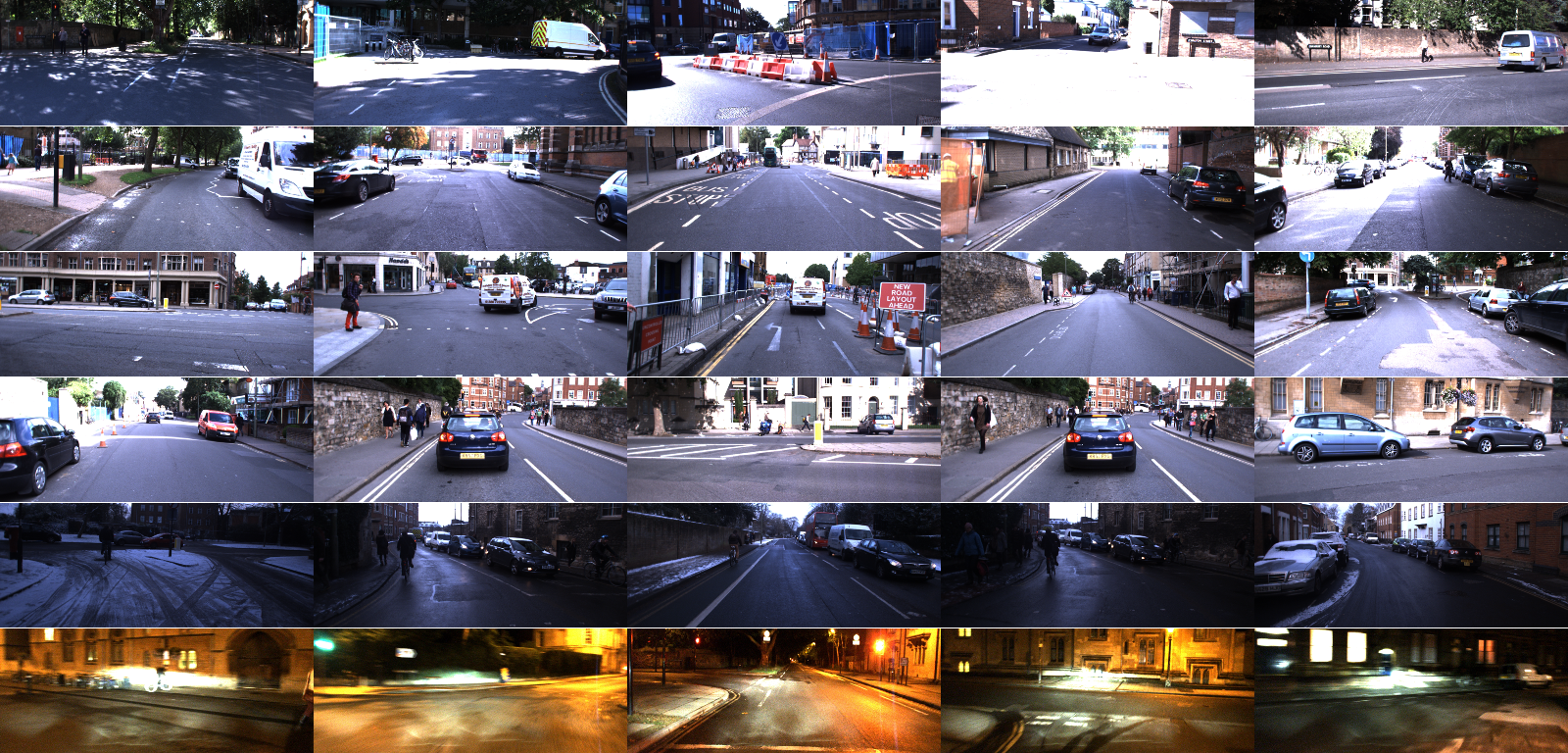}
    \caption{\captionSize Incremental changes of lighting conditions in the Oxford10k dataset from early morning (top row) to late night (bottom row). }
	\label{fig:oxford}
\end{figure}

We evaluate \met~in this context for a drivable-path segmentation method based on segments of the Oxford RobotCar dataset \cite{RobotCarDatasetIJRR}. The employed path segmentation labels are generated in a self-supervised setting based on \cite{BarnesMP16}. The dataset consists of approximately hour-long driving sessions from different days collected over the course of a year. Based on the nature of the dataset, we approximate the scenario of continuous application by picking five datasets to represent different daylight conditions from morning to evening and train on a labelled source dataset based on morning data as seen in Figure \ref{fig:oxford}. 

The resulting 5 intermediate domains were chosen to represent incremental change in lighting conditions and serve as a proxy for the online deployment scenario. Each domain consists of about 2000 images, rescaled for the evaluation to a size of 128 by 320 pixels. Pixel-wise segmentation labels for training are available only for the source domain, while the approach utilises test labels for the evaluation in all domains.

For all segmentation tasks, we employ an adaptation of the ENet~\cite{paszke2016enet} architecture which presents a compromise of model performance and size. The architecture focuses on strong segmentation accuracy as well as reasonable computational requirements, which makes it a strong contender for online deployment on mobile platforms. For the discriminator, we split the ENet architecture just before the upsampling stages (see \cite{paszke2016enet}) and employ an additional 4-layer convolutional discriminator. Similar to Section \ref{sec:mnist} we duplicate all parameters before the architecture split to be utilised as source and target encoders. 

The results for drivable-path segmentation are represented in Table \ref{tab:segmentation}. 
Similar to the application on the synthetic domain shift dataset in Section \ref{sec:mnist}, \met~outperforms one-step domain adaptation. ADA with respect to the union over all incremental target domains is more accurate than with only the final domain but not as exact as \met. Again, \metadd~slightly reduces the target performances, however rendering the storage of significant source datasets unnecessary.

In real-world scenarios, we cannot ensure smoothness over the appearance changes and the turning-on of street lights for the final target domain indeed represents a step change in our environment. It is to be expected that more continuous domain shifts would increase the advantages of \met~as displayed in the context of synthetic data in Section \ref{sec:mnist}.

\begin{table}[h]
    \centering
    \begin{tabular}{c|c|c|c|c|c|c}
       \toprule
           \makecell{target \\domains} & \makecell{only\\ source} & ADA & \makecell{ADA \\\metadd} &\makecell{ADA\\Union} & \met & \makecell{\met~\\\metadd} \\
        \midrule
         morning & 91.62 & - & - & - & 91.60 & 91.77 \\
         midday & 90.70 & - & - & - & 91.05 & 90.50 \\
         afternoon & 89.10 & - & - & - & 89.91 & 89.53 \\
         evening & 87.08 & - & - & - & 89.01 & 87.34 \\
         night & 76.27 & 78.67 & 77.12 & 78.83 & 80.21 & 79.37 \\
        \bottomrule
    \end{tabular}
    \vspace{3pt}
    \caption{\captionSize Mean average precision results for segmentation task in continuous deployment scenario. Applying domain adaptation with respect to the union of all target domains slightly increases performance. The incremental adaptation approach leads to further improvement, while the approximation of the source domain only slightly reduces performance.}
    \label{tab:segmentation}
\end{table}

In comparison to the toy example, the combination of all target domains only leads to minor improvement over the regular application of ADA with only the final target domain. 
The final target domain's instantaneous change in lighting due to the switching-on of the street lights leads to significant differences to previous target domains. This scenario renders domain adaptation to the union over all target domains less efficient and emphasising the importance of an incremental method, which focuses on the current target domain. 

With computation times of approximately 26 minutes for the adaptation to a new incremental target domain on an NVIDIA GeForce GTX Titan Xp GPU, we can potentially deploy the system on vehicles to adapt to the currently encountered domain at a rate of about 55 times a day in continuous deployment. The extension with source domain modelling reaches computation times of 29 minutes resulting in nearly 50 updates per day. 

\begin{figure*}[tb]
	\centering
		\includegraphics[width = 1\textwidth]{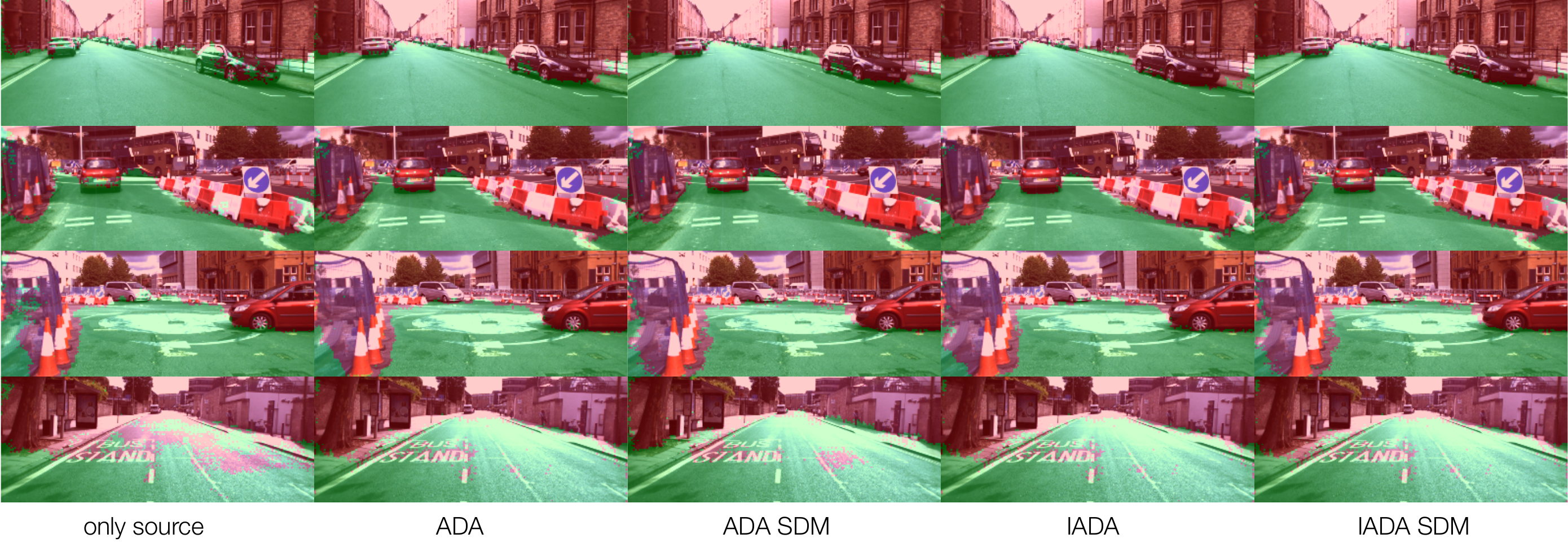}
    \caption{\captionSize Segmentation predictions for the final target domain overlayed on the input images (green and red represent traversable terrain and obstacles respectively). From left to right: training only on the source domain, ADA, \met, ADA \metadd, \met~\metadd. Adversarial domain adaptation consistently outperforms source training, with \met~providing additional benefits. When combined with \metadd~both approaches result in only slightly lower accuracy. While only source domain trained models display obvious weaknesses correlated to the different street illumination, the main benefits of \met~against ADA can be found in details such as more distinct obstacle boundaries and less noisy segmentation. The slight performance reduction based on \metadd~is qualitatively negligible and mostly visible in the quantitative evaluation in Table \ref{tab:segmentation}}
	\label{fig:oxford_output}

\end{figure*}

\section{Discussion}
\label{sec:discussion}

While \met's principal benefits are based on continuous access to the incremental shifts between source and target domains, the evaluation for drivable-path segmentation with our offline datasets builds on a sequence of distinct target domains extracted from the Oxford RobotCar Dataset. 
The approach can be extended easily to more continuous alignment to the online perceived data domain via the utilisation of sliding window sampling during deployment. 
Interestingly, it was shown in Section \ref{sec:mnist} that the benefits of dividing the target domains further for \met~can saturate when the intermediate domains are becoming increasingly similar.

\met~relies on access to the incremental shifts in the appearance of our environment. With limited access or step-wise changes in the perceived environment the approach degrades to regular adversarial domain adaptation. In particular, this paradigm becomes visible in our segmentation datasets where the turning-on of the streetlights leads to an instantaneous change in the appearance of the environment.

However, this instantaneous domain shift caused by the final domain's lighting change further emphasises the benefits of an incremental approach over simply using the union over all target domains for regular domain adaptation. \met~ significantly outperforms this method as it specifically optimises for the currently relevant target domain.

All experimental results noted in our work are based on the confusion loss for domain adaptation \cite{wulfmeier2017addressing}. An adaptation of the Wasserstein GAN framework \cite{arjovsky2017wasserstein} for domain adaptation leads to (on average) slightly more stable training and statistically insignificantly improvement in performance. However, we focused on the confusion loss formulation as, due to the additional critic training rounds required for the WGAN framework, it leads to significantly lower training duration. 

While increased computational effort might not be critical for server-side computation, it can limit applications of embedded systems.
In the context of cloud computing or larger platforms with significant data storage volumes, the minor accuracy loss can be prevented when applying the original formulation for \met~in Section \ref{sec:problem}.


\section{Conclusion and Future Work}
\label{sec:conclusion}

We present a method for addressing the task of domain adaptation in an incremental fashion, adapting to a continuous stream of changing target domains. Furthermore, we introduce an approach for source domain modelling, training a GAN to approximate the feature distribution in the source domain to render the domain adaptation  step independent of retaining large amounts of source data.
Both methods are evaluated first on synthetically shifted versions of rescaled MNIST digits for illustration purposes and full access to the number of intermediate domains. Furthermore, we empirically demonstrate their performance on the real-world task of drivable-path segmentation in the context of autonomous driving.

The field of continual training during deployment provides many possible benefits as models can be adapted to the currently encountered environment and learn from data unavailable during offline training. However, the approach also opens up new security challenges. The well-known problem of perpetrators introducing adversarial samples to the system could lead to not only corruption of the current prediction but prolonged distortion of the model. This area represents an essential direction for further research on defending against adversarial examples.
Further indispensable extensions of this work include addressing the additional problem of catastrophic forgetting in lifelong-learning scenarios. This direction has the potential to further reduce computational requirements as it will discard the necessity to readapt to once encountered target domains.

\vspace{2mm}
\section*{Acknowledgment}
The authors would like to acknowledge the support of the UK’s Engineering and Physical Sciences Research Council (EPSRC) through the Programme Grant EP/M019918/1 and the Doctoral Training Award (DTA) as well as the support of the Hans-Lenze-Foundation. Additionally, the donation from NVIDIA of the Titan Xp GPU used in this work is gratefully acknowledged.

\bibliographystyle{unsrt}
\bibliography{main}

\begin{thebibliography}{10}

\bibitem{ratnasingam2010study}
Sivalogeswaran Ratnasingam and Steve Collins.
\newblock Study of the photodetector characteristics of a camera for color
  constancy in natural scenes.
\newblock {\em JOSA A}, 27(2):286--294, 2010.

\bibitem{Long0J16a}
Mingsheng Long, Jianmin Wang, and Michael~I. Jordan.
\newblock Deep transfer learning with joint adaptation networks.
\newblock {\em CoRR}, abs/1605.06636, 2016.

\bibitem{Ganin2016}
Yaroslav Ganin, Evgeniya Ustinova, Hana Ajakan, Pascal Germain, Hugo
  Larochelle, Fran{\c{c}}ois Laviolette, Mario Marchand, Victor Lempitsky, Urun
  Dogan, Marius Kloft, Francesco Orabona, and Tatiana Tommasi.
\newblock {Domain-Adversarial Training of Neural Networks}.
\newblock {\em Journal of Machine Learning Research}, 17:1--35, 2016.

\bibitem{ajakan2014domain}
Hana Ajakan, Pascal Germain, Hugo Larochelle, Fran{\c{c}}ois Laviolette, and
  Mario Marchand.
\newblock Domain-adversarial neural networks.
\newblock {\em arXiv preprint arXiv:1412.4446}, 2014.

\bibitem{wulfmeier2017addressing}
Markus Wulfmeier, Alex Bewley, and Ingmar Posner.
\newblock Addressing appearance change in outdoor robotics with adversarial
  domain adaptation.
\newblock In {\em Proceedings of the IEEE International Conference on
  Intelligent Robots and Systems}, 2017.

\bibitem{bousmalis2016domain}
Konstantinos Bousmalis, George Trigeorgis, Nathan Silberman, Dilip Krishnan,
  and Dumitru Erhan.
\newblock Domain separation networks.
\newblock In {\em Advances in Neural Information Processing Systems}, pages
  343--351, 2016.

\bibitem{tzeng2015towards}
Eric Tzeng, Coline Devin, Judy Hoffman, Chelsea Finn, Xingchao Peng, Sergey
  Levine, Kate Saenko, and Trevor Darrell.
\newblock Towards adapting deep visuomotor representations from simulated to
  real environments.
\newblock {\em arXiv preprint arXiv:1511.07111}, 2015.

\bibitem{tzeng2017adversarial}
Eric Tzeng, Judy Hoffman, Kate Saenko, and Trevor Darrell.
\newblock Adversarial discriminative domain adaptation.
\newblock {\em arXiv preprint arXiv:1702.05464}, 2017.

\bibitem{hoffman2016fcns}
Judy Hoffman, Dequan Wang, Fisher Yu, and Trevor Darrell.
\newblock Fcns in the wild: Pixel-level adversarial and constraint-based
  adaptation.
\newblock {\em arXiv preprint arXiv:1612.02649}, 2016.

\bibitem{ShrivastavaPTSW16}
Ashish Shrivastava, Tomas Pfister, Oncel Tuzel, Josh Susskind, Wenda Wang, and
  Russ Webb.
\newblock Learning from simulated and unsupervised images through adversarial
  training.
\newblock {\em CoRR}, abs/1612.07828, 2016.

\bibitem{goodfellow2014generative}
Ian Goodfellow, Jean Pouget-Abadie, Mehdi Mirza, Bing Xu, David Warde-Farley,
  Sherjil Ozair, Aaron Courville, and Yoshua Bengio.
\newblock Generative adversarial nets.
\newblock In {\em Advances in neural information processing systems}, pages
  2672--2680, 2014.

\bibitem{RobotCarDatasetIJRR}
Will Maddern, Geoff Pascoe, Chris Linegar, and Paul Newman.
\newblock {1 Year, 1000km: The Oxford RobotCar Dataset}.
\newblock {\em The International Journal of Robotics Research (IJRR)},
  36(1):3--15, 2017.

\bibitem{lowry2016visual}
Stephanie Lowry, Niko S{\"u}nderhauf, Paul Newman, John~J Leonard, David Cox,
  Peter Corke, and Michael~J Milford.
\newblock Visual place recognition: A survey.
\newblock {\em IEEE Transactions on Robotics}, 32(1):1--19, 2016.

\bibitem{ChurchillIJRR2013}
Winston Churchill and Paul Newman.
\newblock {E}xperience-based {N}avigation for {L}ong-term {L}ocalisation.
\newblock {\em The International Journal of Robotics Research (IJRR)}, 2013.

\bibitem{neubert2013appearance}
Peer Neubert, Niko Sunderhauf, and Peter Protzel.
\newblock Appearance change prediction for long-term navigation across seasons.
\newblock In {\em Mobile Robots (ECMR), 2013 European Conference on}, pages
  198--203. IEEE, 2013.

\bibitem{Hong2017}
Sungeun Hong, Woobin Im, Jongbin Ryu, and Hyun~S Yang.
\newblock Sspp-dan: Deep domain adaptation network for face recognition with
  single sample per person.
\newblock {\em arXiv preprint arXiv:1702.04069}, 2017.

\bibitem{rozantsev2016beyond}
Artem Rozantsev, Mathieu Salzmann, and Pascal Fua.
\newblock Beyond sharing weights for deep domain adaptation.
\newblock {\em arXiv preprint arXiv:1603.06432}, 2016.

\bibitem{SunFS16}
Baochen Sun, Jiashi Feng, and Kate Saenko.
\newblock Correlation alignment for unsupervised domain adaptation.
\newblock {\em CoRR}, abs/1612.01939, 2016.

\bibitem{arjovsky2017towards}
Martin Arjovsky and L{\'e}on Bottou.
\newblock Towards principled methods for training generative adversarial
  networks.
\newblock In {\em NIPS 2016 Workshop on Adversarial Training. In review for
  ICLR}, volume 2016, 2017.

\bibitem{kembhavi2009incremental}
Aniruddha Kembhavi, Behjat Siddiquie, Roland Miezianko, Scott McCloskey, and
  Larry~S Davis.
\newblock Incremental multiple kernel learning for object recognition.
\newblock In {\em Computer Vision, 2009 IEEE 12th International Conference on},
  pages 638--645. IEEE, 2009.

\bibitem{jain2011online}
Vidit Jain and Erik Learned-Miller.
\newblock Online domain adaptation of a pre-trained cascade of classifiers.
\newblock In {\em Computer Vision and Pattern Recognition (CVPR), 2011 IEEE
  Conference on}, pages 577--584. IEEE, 2011.

\bibitem{bewley2014online}
Alex Bewley, Vitor Guizilini, Fabio Ramos, and Ben Upcroft.
\newblock Online self-supervised multi-instance segmentation of dynamic
  objects.
\newblock In {\em Robotics and Automation (ICRA), 2014 IEEE International
  Conference on}, pages 1296--1303. IEEE, 2014.

\bibitem{hoffman2014continuous}
Judy Hoffman, Trevor Darrell, and Kate Saenko.
\newblock Continuous manifold based adaptation for evolving visual domains.
\newblock In {\em Proceedings of the IEEE Conference on Computer Vision and
  Pattern Recognition}, pages 867--874, 2014.

\bibitem{lin2013network}
Min Lin, Qiang Chen, and Shuicheng Yan.
\newblock Network in network.
\newblock {\em arXiv preprint arXiv:1312.4400}, 2013.

\bibitem{clevert2015fast}
Djork-Arn{\'e} Clevert, Thomas Unterthiner, and Sepp Hochreiter.
\newblock Fast and accurate deep network learning by exponential linear units
  (elus).
\newblock {\em arXiv preprint arXiv:1511.07289}, 2015.

\bibitem{BarnesMP16}
Dan Barnes, William~P. Maddern, and Ingmar Posner.
\newblock Find your own way: Weakly-supervised segmentation of path proposals
  for urban autonomy.
\newblock {\em CoRR}, abs/1610.01238, 2016.

\bibitem{paszke2016enet}
Adam Paszke, Abhishek Chaurasia, Sangpil Kim, and Eugenio Culurciello.
\newblock Enet: A deep neural network architecture for real-time semantic
  segmentation.
\newblock {\em arXiv preprint arXiv:1606.02147}, 2016.

\bibitem{arjovsky2017wasserstein}
Martin Arjovsky, Soumith Chintala, and L{\'e}on Bottou.
\newblock Wasserstein gan.
\newblock {\em arXiv preprint arXiv:1701.07875}, 2017.

\end{thebibliography}

\end{document}